%% file: main.tex
\documentclass[10pt,twocolumn,letterpaper]{article}

\usepackage[pagenumbers]{cvpr} % To force page numbers, e.g. for an arXiv version


\definecolor{cvprblue}{rgb}{0.21,0.49,0.74}
\usepackage[pagebackref,breaklinks,colorlinks,citecolor=cvprblue]{hyperref}

\def\paperID{*****} % *** Enter the Paper ID here
\def\confName{CVPR}
\def\confYear{2024}

\title{3rd Place Solution for MeViS Track in CVPR 2024 PVUW workshop: Motion Expression guided Video Segmentation}

\author{Feiyu Pan\thanks{Equal contribution} \hspace{1cm} Hao Fang\footnotemark[1] \hspace{1cm} Xiankai Lu\thanks{Corresponding author} \\
TIME Team, School of Software, Shandong University \\
{\tt\small \{panfeiyu,fanghaook\}@mail.sdu.edu.cn, carrierlxk@gmail.com}
}

\begin{document}
\maketitle
\begin{abstract}
Referring video object segmentation (RVOS) relies on natural language expressions to segment target objects in video, emphasizing modeling dense text-video relations. The current RVOS methods typically use independently pre-trained vision and language models as backbones, resulting in a significant domain gap between video and text. In cross-modal feature interaction, text features are only used as query initialization and do not fully utilize important information in the text. In this work, we propose using frozen pre-trained vision-language models (VLM) as backbones, with a specific emphasis on enhancing cross-modal feature interaction. Firstly, we use frozen convolutional CLIP backbone to generate feature-aligned vision and text features, alleviating the issue of domain gap and reducing training costs. Secondly, we add more cross-modal feature fusion in the pipeline to enhance the utilization of multi-modal information. Furthermore, we propose a novel video query initialization method to generate higher quality video queries. Without bells and whistles, our method achieved 51.5 \( \mathcal{J} \)\&\( \mathcal{F} \) on the MeViS test set and ranked 3rd place for MeViS Track in CVPR 2024 PVUW workshop: Motion Expression guided Video Segmentation.
\end{abstract}

\section{Introduction}
\label{sec:intro}
Referring video object segmentation (RVOS) is a continually evolving task that aims to segment target objects in video, referred to by linguistic expressions. To investigate the feasibility of using motion expressions to ground and segment objects in videos,  a large-scale dataset called MeViS~\cite{ding2023mevis} was proposed, which contains a large number of motion expressions to indicate target objects in complex environments. Therefore, Motion Expression guided Video Segmentation requires segmenting objects in video content based on sentences describing object motion, which is more challenging compared to traditional RVOS datasets.

\begin{figure}[t]
\begin{center}
\includegraphics[width=1\linewidth]{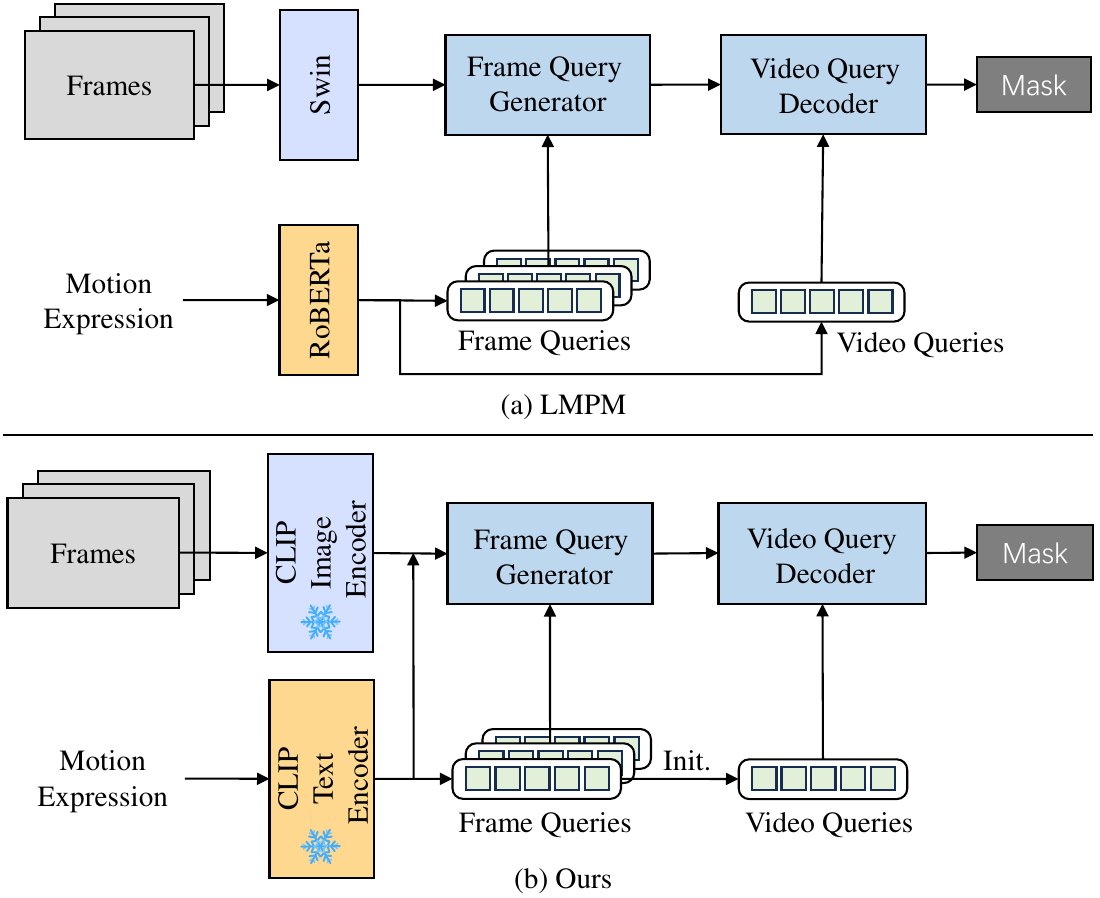}
\end{center}
   \vspace{-5mm}
\caption{Comparisons of LMPM and our Framework. 
\vspace{-5mm}}
\label{fig:intro}
\end{figure}

The MeViS dataset further emphasizes the importance of language understanding and modeling text-video relations. The current RVOS methods~\cite{botach2022end,wu2022language,ding2023mevis,li2023robust,han2023html,wu2023onlinerefer} typically use independently pre-trained vision and language models as backbones. As shown in Fig.~\ref{fig:intro} (a), LMPM~\cite{ding2023mevis} uses Tiny Swin Transformer~\cite{liu2021swin} as the image encoder and RoBERTa~\cite{liu2019roberta} as the text encoder. This leads to a significant domain gap between video and text, making text-video relation modeling more difficult and requiring more training costs to finetune the backbone. In addition, previous methods do not place enough emphasis on cross-modal feature interaction. For example, LMPM~\cite{ding2023mevis} only use text embedding as query initialization and do not fully utilize key information in the text. 

\begin{figure*}[t]
\begin{center}
\includegraphics[width=1\linewidth]{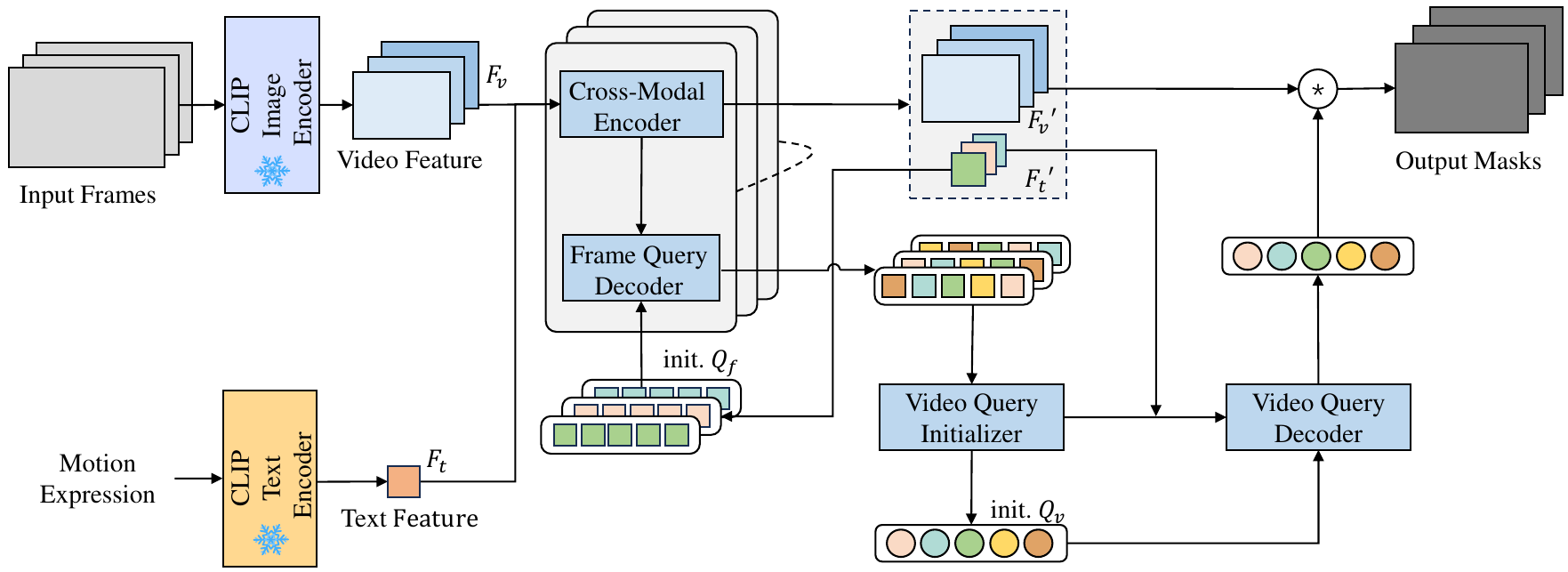}
\end{center}
   \vspace{-3mm}
\caption{The overview architecture of the proposed method. This model inputs multiple images and motion expression from video clips, and outputs multi-scale visual features and sentence-level text features through a frozen CLIP backbone. Cross-Modal Encoder fuses text and image features, Frame Query decoder independently generates frame queries for each frame. Then, Video Query Decoder reorder and fuses all frame queries to adaptively initialize video queries. Finally, Video Query Decoder refines video queries for final mask prediction.}
\label{fig:model}
\end{figure*}

In this work, we propose using frozen pre-trained vision-language models (VLM) as backbones, with a specific emphasis on enhancing cross-modal feature interaction. Firstly, we use frozen convolutional CLIP~\cite{liu2022convnet,radford2021learning} backbone to generate feature-aligned vision and text features. As shown in Fig.~\ref{fig:intro} (b), we do not fine tune the CLIP backbone to preserve pre-trained knowledge of vision-language association. This not only alleviates the issue of domain gap, but also greatly reduces training costs. Secondly, we add more cross-modal feature fusion in the pipeline to enhance the utilization of multi-modal information. We design three cross-modal feature interaction module in the model, including cross-modal encoder, frame query decoder and video query decoder. These modules enhance video and text features through simple cross-attention. Furthermore, to fully utilize the prior knowledge of frame queries, we propose a novel video query initialization method to generate higher quality video queries. Specifically, we perform bipartite matching and reorder frame queries, then aggregate them in a weighted manner to initialize video queries.

In this year, Pixel-level Video Understanding in the Wild Challenge (PVUW) challenge adds two new tracks, Complex Video Object Segmentation Track based on MOSE~\cite{ding2023mose} and Motion Expression guided Video Segmentation track based on MeViS~\cite{ding2023mevis}. In the two new tracks, additional videos and annotations that feature challenging elements are provided, such as the disappearance and reappearance of objects, inconspicuous small objects, heavy occlusions, and crowded environments in MOSE~\cite{ding2023mose}. Moreover, a new motion expression guided video segmentation dataset MeViS~\cite{ding2023mevis} is provided to study the natural language-guided video understanding in complex environments. These new videos, sentences, and annotations enable us to foster the development of a more comprehensive and robust pixel-level understanding of video scenes in complex environments and realistic scenarios. We make evaluations of the proposed method on the MeViS~\cite{ding2023mevis} dataset. Without using any additional training data, our method achieved 46.9 \( \mathcal{J} \)\&\( \mathcal{F} \) on the MeViS val set, 51.5 \( \mathcal{J} \)\&\( \mathcal{F} \) on the MeViS test set and ranked 3rd place for MeViS Track in CVPR 2024 PVUW workshop: Motion Expression guided Video Segmentation.

% \section{Related Work}
% \label{sec:relat}
% \noindent \textbf{Referring Video Object Segmentation.} 

% \noindent \textbf{Vision-Language Models.} 

\begin{figure*}[t]
\begin{center}
\includegraphics[width=1\linewidth]{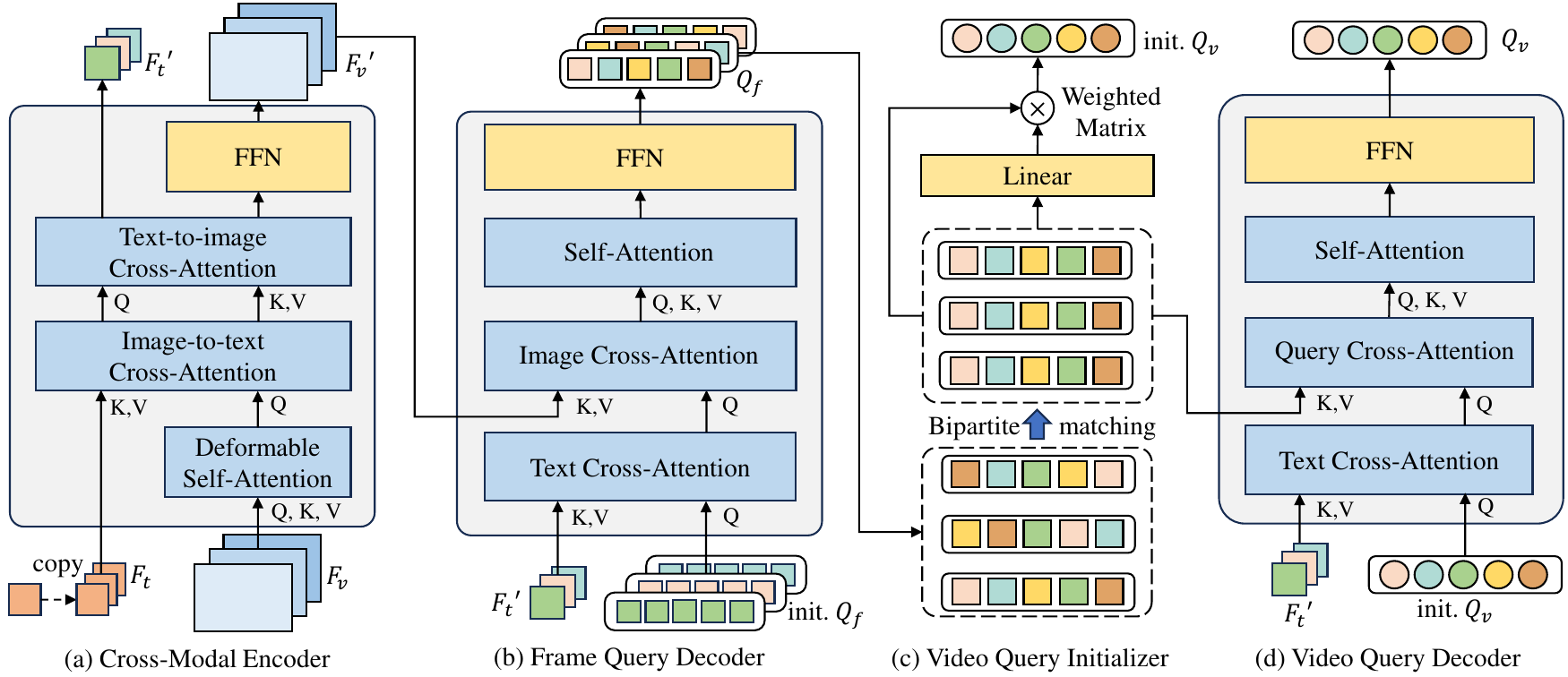}
\end{center}
   \vspace{-5mm}
\caption{Illustration of Cross-modal Encoder, Frame Query Decoder, Video Query Initializer and Video Query Decoder.
\vspace{-2mm}}
\label{fig:module}
\end{figure*}

\section{Method}
\label{sec:metho}
We propose a novel framework for referring video object segmentation. It contains a frozen convolutional CLIP image backbone for video feature extraction, a CLIP text backbone for text feature extraction, a cross-modal encoder for image and text feature fusion (Sec.~\ref{sec:encoder}), a frame query decoder for generating frame queries (Sec.~\ref{sec:frame decoder}), a video query initializer for video query initialization (Sec.~\ref{sec:video init}), and a video query decoder for mask refinement (Sec.~\ref{sec:video decoder}). The overall framework is available in Fig.~\ref{fig:model}.

\subsection{Cross-modal Encoder}
\label{sec:encoder}
Given an \texttt{(Video, Text)} pair, we extract multi-frame multi-scale image features $F_v$ with CLIP image encoder, and text features $F_t$ with CLIP test encoder. Due to the use of convolutional CLIP image encoder~\cite{yu2024convolutions}, we can extract multi-scale features from the outputs of different blocks. After extracting vanilla video and text features, we fed them into a cross-modal encoder for cross-modal feature fusion. As shown in Fig.~\ref{fig:module} (a), the cross-modal encoder is built on top of the pixel decoder of Mask2Former~\cite{cheng2022masked}, which leverages the Deformable self-attention~\cite{zhu2020deformable} to enhance image features. Inspired by Grounding DINO~\cite{liu2023grounding} and GLIP~\cite{li2022grounded}, we add an image-to-text cross-attention and a text-to-image cross-attention for feature fusion. These modules help align features of different modalities, ultimately obtaining enhanced image features $F^{'}_{v}$ and text features $F^{'}_{t}$.

\subsection{Frame Query Decoder}
\label{sec:frame decoder}
We develop a frame query decoder to independently generate frame queries $Q_{f} \in \mathbb{R}^{T \times N_{f}\times C}$ for each frame, as shown in Fig.~\ref{fig:module} (b). Frame queries are directly initialized by text features, then are fed into a text cross-attention layer to combine text features, an image cross-attention layer to combine image features, a self-attention layer, and an FFN layer in each frame query decoder layer. Each decoder layer has an extra text cross-attention layer compared with the transformer decoder layer of Mask2Former~\cite{cheng2022masked}, as we need to inject text information into queries for better modality alignment.

\subsection{Video Query Initializer}
\label{sec:video init}
After generating frame-level representation, the next step is to generate video queries  $Q_{v} \in \mathbb{R}^{N_{v}\times C}$ to represent the entire video clip. Inspired by LBVQ~\cite{fang2024learning}, video queries have great similarity to frame queries per frame, and their essence is the fusion of frame queries. Instead of the simple text feature initialization strategy~\cite{ding2023mevis}, we aggregate frame queries to achieve video query initialization. 

Firstly, as shown in Fig.~\ref{fig:module} (c), the Hungarian matching algorithm~\cite{kuhn1955hungarian} is utilized to match the $Q_f$ of adjacent frames, as is done in ~\cite{huang2022minvis}.
\begin{equation}
  \left\{
  \begin{aligned}
  \tilde{Q}_{f}^{t} & = \text{Hungarian}(Q_{f}^{t-1},Q_{f}^{t}), \quad t\in[2,T] \\
   \tilde{Q}_{f}^{t} & = Q_{f}^{t}, \quad t=1
  \end{aligned}
  \right.
\label{equ:match},
\end{equation}
where $\tilde{Q}_{f}$ represents rearranged frame queries. The purpose of this operation is to ensure that the instance order of each frame query is consistent. Then, due to the varying importance of each frame, we aggregate frame queries using learnable weights.
\begin{equation}
  \label{equ:AQI}
  Q_{v} = \sum_{t=1}^{T}\text{Softmax}(\text{Linear}(\tilde{Q}_{f}^{t}))\tilde{Q}_{f}^{t},
\end{equation}
where $\text{Linear}$ denotes a learnable weight that is realized by linear layer with input dimension $C$ and output dimension $1$. The weights of different frames are maintained as a sum of 1 through the \text{Softmax} function.

\subsection{Video Query Decoder}
\label{sec:video decoder}
After obtaining the initialized video queries, they are fed into the video query decoder for layer by layer refinement. As shown in Fig.~\ref{fig:module} (d), video queries are fed into a text cross-attention layer to combine text features, an query cross-attention layer to combine frame queries features, a self-attention layer, and an FFN layer in each video query decoder layer. The video queries of the last layer will be dot multiplied with image features to generate the final mask.

\subsection{Training Loss}
Following~\cite{heo2022vita,ding2023mevis}, we employ the match loss $\mathcal{L}_{f}$ between per-frame outputs and frame-wise ground truth, along with $\mathcal{L}_{v}$ as the video-level loss with video-level ground truth. The total training objective to optimize the model is: $\mathcal{L}_{train} = \mathcal{L}_{v} + \mathcal{L}_{f} + \lambda_{sim}\mathcal{L}_{sim}$, where $\lambda_{sim}$ is used for balancing the similarity loss $\mathcal{L}_{sim}$.

\section{Experiment}
\label{sec:exper}
\subsection{Datasets and Metrics}
\noindent \textbf{Datasets.} 
MeViS~\cite{ding2023mevis} is a newly established dataset that is targeted at motion information analysis and contains 2,006 video clips and 443k high-quality object segmentation masks, with 28,570 sentences indicating 8,171 objects in complex environments. All videos are divided into 1,662 training videos, 190 validation videos and 154 test videos.

\noindent \textbf{Evaluation Metrics.} 
we employ region similarity $\mathcal{J}$ (average IoU), contour accuracy $\mathcal{F}$ (mean boundary similarity), and their average \( \mathcal{J} \)\&\( \mathcal{F} \) as our evaluation metrics.

\subsection{Implementation Details}
We use ConvNeXt-Large CLIP~\cite{liu2022convnet,radford2021learning} backbones from OpenCLIP~\cite{openclip} pretrained on LAION-2B~\cite{schuhmann2022laion} dataset. On top of the frozen CLIP backbone, we build our model following Mask2Former~\cite{cheng2022masked}. By default, Cross-modal Encoder is composed of six layers, Frame Query Decoder employs nine layers with $N_f$ = 20 frame queries, and Video Query Decoder employs six layers with $N_v$ = 20 video queries. The coefficients for similarity loss is set as $\lambda_{sim}$ = 0.5. We train 100,000 iterations using AdamW optimizer~\cite{loshchilov2017decoupled} with a learning rate of 0.00005. Our model is trained on 4 NVIDIA 3090 GPUs, each with a video clip containing 8 randomly selected frames. All frames are cropped to have the longest side of 640 pixels and the shortest side of 360 pixels during training and evaluation.

\subsection{Main Results}
As shown in Tab.~\ref{tab:leader}, our approach achieves 51.5\% in \( \mathcal{J} \)\&\( \mathcal{F} \), ranking in 3rd place for MeViS Track in CVPR 2024 PVUW workshop: Motion Expression guided Video Segmentation.

\begin{table}
  \centering
  \begin{tabular}{l|ccc}
    \toprule
    Team & \( \mathcal{J} \)\&\( \mathcal{F} \) & $\mathcal{J}$ & $\mathcal{F}$ \\
    \midrule
    Tapallai & 54.5 & 50.5 & 58.5\\
    BBBiiinnn & 54.2 & 51.0 & 57.4\\
    \bf Ours & 51.5 & 46.1 & 56.9\\
    LIULINKAI & 42.7 & 39.3 & 46.1\\
    ntuLC & 37.0 & 34.1 & 39.9\\
  \bottomrule
\end{tabular}
\caption{The leaderboard of the MeViS test set.}
\label{tab:leader}
\end{table}

\subsection{Ablation Study}
We conducted ablation experiments on the CLIP backbone. As shown in Tab.~\ref{tab:ablation}, ConvNeXt-L backbone showed a 2.7\% \( \mathcal{J} \)\&\( \mathcal{F} \) performance improvement compared to ResNet50 backbone.
\begin{table}
  \centering
  \begin{tabular}{l|ccc}
    \toprule
    Backbone & \( \mathcal{J} \)\&\( \mathcal{F} \) & $\mathcal{J}$ & $\mathcal{F}$ \\
    \midrule
    ResNet50 & 44.2 & 39.6 & 48.8\\
    ConvNeXt-L & 46.9 & 41.9 & 51.8\\
  \bottomrule
\end{tabular}
\caption{Ablation study of backbone on MeViS validation set.}
\label{tab:ablation}
\end{table}

\section{Conclusion}
In this work, we propose using frozen pre-trained vision-language models as backbones, with a specific emphasis on enhancing cross-modal feature interaction. We use frozen convolutional CLIP backbone to generate feature-aligned vision and text features, alleviating the issue of domain gap and reducing training costs. We add more cross-modal feature fusion in the pipeline to enhance the utilization of multi-modal information. We propose a novel video query initialization method to generate higher quality video queries. Evaluations are made on the MeViS dataset and our method ranked 3rd place for MeViS Track in CVPR 2024 PVUW workshop: Motion Expression guided Video Segmentation.

{
    \small
    \bibliographystyle{ieeenat_fullname}
    \bibliography{main}
}

\end{document}